\documentclass[11pt,a4paper]{article}
\usepackage[hyperref]{acl2020}
\usepackage{times}
\usepackage{latexsym}
\usepackage{pgfplots}
\usepackage[LGR,T1]{fontenc}
\usepackage[utf8]{inputenc}
\usepackage{multirow}

\usepackage{microtype}

\aclfinalcopy

\newcommand{\textgreek}[1]{\begingroup\fontencoding{LGR}\selectfont#1\endgroup}

\title{Automatically Ranked Russian Paraphrase Corpus for Text Generation}

\author{
  Vadim Gudkov$^1$,  Olga Mitrofanova$^2$,  Elizaveta Filippskikh$^3$\\\\
  Saint Petersburg State University$^1$$^,$$^2$ \\
  \texttt{st071220@student.spbu.ru}$^1$ , \texttt{o.mitrofanova@spbu.ru}$^2$ \\
  CraftTalk$^3$ \\
  \texttt{efilippskikh@crafttalk.ru}$^3$ \\
}

\date{}

\begin{document}
\maketitle
\begin{abstract}
The article is focused on automatic development and ranking of a large corpus for Russian paraphrase generation which proves to be the first corpus of such type in Russian computational linguistics. Existing manually annotated paraphrase datasets for Russian are limited to small-sized ParaPhraser corpus and ParaPlag which are suitable for a set of NLP tasks, such as paraphrase and plagiarism detection, sentence similarity and relatedness estimation, etc. Due to size restrictions, these datasets can hardly be applied in end-to-end text generation solutions. Meanwhile, paraphrase generation requires a large amount of training data. In our study we propose a solution to the problem: we collect, rank and evaluate a new publicly available headline paraphrase corpus (ParaPhraser Plus), and then perform text generation experiments with manual evaluation on automatically ranked corpora using the Universal Transformer architecture.
\end{abstract}

\section{Introduction}

A large amount of work is dedicated for a clear understanding of the nature of a paraphrase. On the one hand, traditional theories of language allow to trace the
notion of paraphrase back to the ancient rhetorical tradition (cf. Greek
\textgreek{παράφρασις} ‘retelling’) and treat it quite broadly in case of different types of prose, verse, musical pieces, etc. On the other hand, the generative trend in linguistic research encouraged description of transformations involved in the transition from deep to surface structures and at the same time responsible for the emergence of a wide range of paraphrases, cf. Chomskian generative grammar giving account of various lexical transformations, Melchuk’s Sense-Text theory postulating the process of paraphrasing as synonymic
conversion, etc. In recent works paraphrases are treated as “alternative expressions of the same (or similar) meaning” \citep{agirre-etal-2015-semeval}. Ranking paraphrases as regards their similarity in form and meaning is reflected in a set of paraphrase classifications, where precise paraphrases are distinguished from quasi-paraphrases and non-paraphrases \citep{andrew2007scalable}. At the same time, paraphrase corpora development required deep analysis of paraphrase transformations types (e.g. morphosyntactic, lexical and semantic shifts).

Paraphrasing plays an important role in a broad range of NLP tasks, including but not limited to question answering, summarization, information retrieval, sentence simplification, machine translation and dialogue systems. However, in order to be able to train a good paraphrasing system, large parallel corpora are required, which can be a problem in underdeveloped languages from a data resources standpoint. In order to bridge this gap, we propose a methodology to collect enough data for proper deep learning.

\section{Motivation and Related Work}

Paraphrase identification inspired a set of NLP competitions within SemEval
conferences in 2012, 2013, 2015 and 2016, so that baseline decisions and their
improvements were worked out for English. There also exist several well-known manually annotated paraphrase datasets for English: Microsoft Paraphrase
\citep{dolan2005automatically}, Quora Question Pairs and MS COCO \citep{lin2014microsoft}.

However, Russian is less represented in paraphrase research both in case of resource
development and algorithm evaluation, a few exceptions being AINL Paraphrase detection competition in 2016 based on Paraphraser corpus and
Dialogue Paraphrased plagiarism detection competition in 2017 based on
ParaPlag corpus. Alongside with Paraphraser and ParaPlag, there are some
paraphrase resources which include Russian language, for instance by Opusparcus \citep{creutz2018open} and PPDB \citep{ganitkevitch-etal-2013-ppdb}.

In our study we mainly focus on the collection, evaluation and generation of the, so called, sentential paraphrases. This approach is different from the collection of PPDB, where sub-sentential paraphrases, such as individual word-pairs, were also included and ParaPlag with main focus on text-level rephrasing.

\begin{table*}[]
\noindent
\centering
\begin{tabular}{p{8cm}ll}
\hline \textbf{Model} & \textbf{F1} & \textbf{Accuracy} \\ \hline
Shallow Neural Networks \citep{pivovarova2017paraphraser} & 79.82 & 76.65 \\ \hline
Linguistic Features Classifier \citep{pivovarova2017paraphraser}   & 81.10 & 77.39  \\ \hline
Machine Translation Based Semantic Similarity \citep{kravchenko2018}  & 78.51 & 81.41 \\ \hline
RuBERT \citep{kuratov2019adaptation} & 87.73 & 84.99 \\ \hline
\end{tabular}
\caption{\label{par-detect-table} Paraphrase detection algorithms evaluation.}
\end{table*}

Recent work \citep{gupta2018deep, fu2019paraphrase, egonmwan2019} provides solid evidence in favour of paraphrase generation by means of seq-2-seq architectures. The main problem, however, is that such systems require significant expansion of existing datasets for proper machine learning \citep{roy-grangier-2019-unsupervised}. The lack of data still remains the greatest obstacle to the development of a stable generation system which could be lexically rich and insensitive to rare words. E.g., the largest datasets supplied with proper annotation seldom exceed 100K samples in size. The authors of the aforementioned articles claim that any user generated content is valuable even though noisy to a certain extent. We propose a solution which overcomes the given problem, and it is based on the denoising procedure which has recently attracted growing attention.
We argue that automatically matched and ranked datasets can be used for paraphrase generation task, especially in low-resource languages, by providing experimental results obtained on the Russian Opusparcus subcorpus and on the novel ParaPhraser Plus corpus.

\section{Source Data}

The ParaPhraser Plus corpus \footnote{Available at:\\ \url{http://paraphraser.ru/download/}} is distilled from a database of news headlines, that was kindly provided by the Russian Internet monitoring service, "Webground". Although, the contents of the resources are pretty similar, the data itself in the original ParaPhraser corpus and the ParaPhraser Plus corpus as well as the methodology used to collect the headlines are not the same by any means.  It is important to note, however, that ranking model which will be described in the corresponding section was based on the original corpus. 
The headlines in "Webground" were initially clustered by events over a ten year span, beginning from
the year 2009. Following the hypothesis that within such theme-based
user-generated clusters the chance of seeing a paraphrase is particularly high,
we formed sets of pairs of all possible combinations within each of them. After
weeding out pairs, consisting of the same tokens, we were left with just over 56
million pairs of potential paraphrases. We have also discarded over 200 thousand
headlines where it was not possible to verify the authorship.

\section{Ranking methodology}

There are several known approaches to paraphrase ranking, including
heuristic scoring \citep{pavlick2015ppdb} and supervised modelling
\citep{creutz2018open}. Heuristic scoring can be effectively conducted in
resources with cross-linguistic support, such as PPDB and Opuspracus. However,
ParaPhraser, as well as our addition, is monolingual, therefore this
approach was not possible. On the other hand, supervised modelling
techniques can be adopted: there is a significant amount of labeled data in the original ParaPhraser corpus  and several approaches to paraphrase identification in Russian headlines have been thoroughly researched and summarized in \citep{pivovarova2017paraphraser}. The methods included shallow neural networks, linguistic features based
classifier and a combination of machine translation with semantic similarity.

However, recent research conducted in \citep{kuratov2019adaptation} shows that deep bidirectional pretrained monolingual transformers improve paraphrase detection in Russian by a large margin. It was shown that finetuning a monolingual BERT based model (RuBERT) on the ParaPhraser corpus yields results far better than all of the aforementioned approaches (see Table ~\ref{par-detect-table}).

The training set in ParaPhraser includes 7,227 pairs of sentences, which are classified by humans into three classes:
2,582 non-paraphrases, 2,957 near-paraphrases,and 1,688 precise-paraphrases. The aforementioned RuBERT model was
fine-tuned to a binary classification task: both near-paraphrases and paraphrases were considered to be a single
class. Such approach helps in automatic ranking: it is possible to sort the items in accordance to the probability of
the paraphrase class in descending order.
The fine-tuned RuBERT model is available as part of the DeepPavlov library \citep{burtsev2018deeppavlov}, which enabled us to adopt this approach in our corpus construction study. 

\section{Ranking evaluation}

In order to evaluate our supervised automatic ranking approach we randomly  select a subsample of 500 pairs for manual annotation. To provide a more thorough comparison analysis we step aside from the original 3-way annotation scheme utilized in  ParaPhraser and adopt the approach provided in \citep{creutz2018open} with more similarity degrees. 

The annotation scheme from the original paper is provided in Table ~\ref{annotation-table}.

\begin{table*}[ht!]
\centering
\begin{tabular}{llp{6cm}}
\hline\
\textbf{Category} & \textbf{Description} & \textbf{Examples}\\
\hline\\
Good (4) & \parbox{6cm}{The two sentences can be used in  the same situation and essentially “mean the same thing”} & \parbox{6cm}{It was a last minute thing <-> This wasn’t planned;  \\ I have goose flesh <-> The hair’s standing upon my arms}\\\\
\hline\\
Mostly Good (3) & \parbox{6cm}{It is acceptable to think that the two sentences refer to the same thing, although one sentence might be more specific than the other one, or there are differences in style.} & \parbox{6cm}{Go to your bedroom <-> Just go to sleep; \\ Next man, move it <-> Next, please; \\ Calvin, now what? <-> What are we doing?}\\\\
\hline\\

Mostly Bad (2) & \parbox{6cm}{There is some connection between the sentences that explains why they occur together, but one would not really consider them to mean the same thing.} & \parbox{6cm}{Did you ask him <-> Have you asked her?; \\ Hello, operator? <-> Yes, operator, I’m trying to get to the police}\\\\
\hline\\
Bad (1) & \parbox{6cm}{There is no obvious connection. \\ The sentences mean different things.} & \parbox{6cm}{She’s over there <-> Take me to him; \\ All the cons <-> Nice and comfy}\\\\
\hline\\
\end{tabular}
\caption{\label{annotation-table} Paraphrase annotation scheme as provided in \citep{creutz2018open}. A pair can also be ranked "in-between" categories (e.g. 2.5 or 3.5).}
\end{table*}

To measure The inter-annotator agreement we use Fleiss Kappa, which
is a Cohen’s Kappa generalization to more than two annotators (in
our case - 5); expected agreement is calculated on the basis of the
assumption that random assignment of categories to items, by any
annotator, is governed by the distribution of items among categories
in the actual world. The annotators reach a fair agreement (Kappa
0.267, p-value < 0.05).

The cosine similarity baseline solution of Word2Vec embeddings
achieves a manual annotation Pearson's correlation coefficient of 0.535. 
Our supervised model rankings for ParaPhraser Plus dramatically improve correlation with human judgments (p = 0.734).

\section{Paraphrase generation}
To test our initial hypothesis we conduct a paraphrase
generation experiment on two datasets: Opusparcus and our
ParaPhraser Plus.

There exist several methods to generate paraphrases. The
following techniques are known: rule-based \citep{McKeown1983FocusCO},
Seq-2-Seq \citep{gupta2018deep, fu2019paraphrase, egonmwan2019,roy-grangier-2019-unsupervised}, reinforcement learning \citep{li2017paraphrase}, deep generative models \citep{iyyer2018adversarial}
and a varied combination \citep{gupta2018deep, mallinson2017paraphrasing} of the later three.

We show the results that can be achieved on large
automatically ranked corpora using a Sequence-to-Sequence model based on the Universal Transformer architecture as it has demonstrated superior performance over the past year in multiple generative tasks, such as abstractive summarization, machine translation and, of course, paraphrase generation. 
\citep{gupta2018deep, mallinson2017paraphrasing, gupta2018deep, fu2019paraphrase, egonmwan2019,roy-grangier-2019-unsupervised}.

As pointed out in \citep{vaswani2017attention},  the
attention heads in the transformer model can be found very
useful in learning grammatical, syntactical, morphological
and semantical behavior in the language, which is essential
in paraphrase generation. Such results are being achieved
thanks to the fact that input vectors are connected to every
other via the attention mechanism, thus allowing the network
to learn complex rephrasing dependencies.
Moreover, contrary to recurrent neural networks, a
transformer can be trained in parallel.

For both datasets, Opusparcus and ParaPhraser Plus, we used the same set of model hyper-parameters: 4 layers in the encoder and decoder with 8 heads of attention. In addition, we added a Dropout of p = 0.3. The models were trained until convergence with the Adam optimizer using a scaled learning rate, as proposed by the authors of the original Transformer

\begin{table*}[]
\noindent
\centering
\begin{tabular}{llllll}
\hline
& \textbf{Metric} & \textbf{250k} & \textbf{500k}   & \textbf{1m} & \textbf{2m}     \\
\hline
\multicolumn{1}{c}{\multirow{2}{*}{\textbf{Opusparcus}}} & BLEU        & 5.04  & 6.54  & 6.58  & 6.46  \\
\multicolumn{1}{c}{}                                     & METEOR      & 28.36 & 30.02 & 31.25 & 33.19 \\
\hline
\multirow{2}{*}{\textbf{ParaPhraser Plus}}                & BLEU & 7.54 & 7.76  & 8.73 & 9.81 \\
                                                         & METEOR & 34.35 & 35.58 & 37.46 & 38.09 \\
\hline

\end{tabular}
\caption{\label{generation-scores-table} Generation scores on the test set of each dataset for different train sizes.}
\end{table*}

We also adopt byte-pair encoding (BPE), a data compression
technique where often occuring pairs of bytes are replaced
by additional extra-alphabet symbols. Thanks to this
approach, the most frequent parts of words are kept in the
vocabulary, while rarely occuring words are replaced by a
sequence of several tokens. Languages with rich morphology
benefit the most as the word endings could be separated
since each word form is definitely less frequent than its
stem. BPE encoding allows us to represent all words,
including the ones unseen during training (e.g. first and
last names, which are common in headlines), with a fixed
vocabular.

\section{Results}
We perform experiments on the above mentioned datasets, and report, both
qualitative and quantitative results of our approach. As can be seen in Table
~\ref{generation-scores-table} which demonstrates the quantitative results, there
is a strong correlation between the size of the training set, selected from top N samples, and the final score
of the model.
We also perform a qualitative analysis by sampling 100 examples of the original phrase, reference and our 2m model generated phrase for human evaluation. We asked 3 annotators to choose their paraphrase preference over three possible options: original paraphrase (Human), generated paraphrase (Machine), no preference (Tie). The results can be seen in Table ~\ref{hum-eval-table}.

\begin{table}
\centering
\begin{tabular}{llll}
\hline \textbf{} & \textbf{Human} & \textbf{Tie} & \textbf{Machine} \\ \hline
Opusparcus & 52.3 & 26.2 & 21.5 \\
ParaPhraser Plus & 60.6 & 23.9 & 14.5 \\
\hline
\end{tabular}
\caption{\label{hum-eval-table} Human evaluation of generated paraphrases.}
\end{table}

For the both corpora, we could see that our model is not reaching human parity yet, having 47.7 and 38.4 of (Machine + Tie) user preference for Opusparcus and ParaPhraser datasets respectively.
Some examples of the produced paraphrases can be seen below (translated into English):

\begin{itemize}
\item \textbf{Original:} \emph{"State Duma may prohibit doctors and teachers from accepting gifts other than flowers"}\\
\textbf{Reference:} \emph{"Teachers and doctors in Russia may be prohibited from accepting gifts"}\\
\textbf{Generated:} \emph{"The State Duma proposed to ban doctors and teachers from accepting gifts"}\\
\vspace*{-0.5cm}
\item \textbf{Original:} \emph{"The Bank of Russia revoked its license from the Yekaterinburg Plateau Bank"} \\
\textbf{Reference:} \emph{"Yekaterinburg Plateau Bank is left without its license"} \\
\textbf{Generated:} \emph{"Central Bank revoked the license from "plateau-bank""} \\
\vspace*{-0.5cm}
\item \textbf{Original:} \emph{"Stocks are ready to rise in the stock market."} \\
\textbf{Reference:} \emph{"Stocks are going to rise on the market"} \\
\textbf{Generated:} \emph{"Stock market ready to go up"} \\
\end{itemize}
\vspace*{-0.5cm}
Despite the fact that both of the training sets are noisy to a certain extent, the model
is able to generalize and generate paraphrases of decent quality (from semantic and grammatical standpoint) for types of content it has never seen during the training phase.

\section{Conclusion}
This study confirms our initial hypothesis that data size restrictions can be effectively resolved with automatically ranked corpora, especially in low-resource languages where large manually annotated datasets are not available.
We also present a newly gathered ParaPhraser Plus corpus and results achieved by a transformer model applied to it.

\section{Future work}
In the future we would like to extend our work to other generative tasks and create more diverse and large ranked corpora utilizing different approaches for supervised ranking. In addition to that, we are interested in investigating how a combination of ranking techniques could be used for better data sampling in generation oriented tasks. Also we would like to investigate what is the minimal amount of manually annotated data that is sufficient for successful automatic ranking in parallel corpora. 

\bibliographystyle{acl_natbib}
\bibliography{acl2020}

\end{document}